\documentclass[11pt,a4paper]{article}
\usepackage[hyperref]{acl2021}
\usepackage{times}
\usepackage{latexsym}

\usepackage{microtype}

\aclfinalcopy 

\usepackage{times}  
\usepackage{helvet} 
\usepackage{courier}  
\usepackage{graphicx} 

\usepackage{booktabs} 
\usepackage{subfigure} 
\usepackage{blindtext}
\usepackage{algorithm} 
\usepackage[noend]{algpseudocode} 
\usepackage{tikz} 
\usepackage{setspace} 
\usepackage{color}
\usepackage{amsmath}
\usepackage{amssymb}
\usepackage{amsthm}

\usepackage{enumerate}
\usepackage{multirow}

\usepackage{esvect}

\usepackage{color,soul}

\title{Encoding Explanatory Knowledge for Zero-shot \\Science Question Answering}

\author{
  Zili Zhou\textsuperscript{1,2}, Marco Valentino\textsuperscript{1}, D\'onal Landers\textsuperscript{2}, Andr\'e Freitas\textsuperscript{1,2,3} \\
  \textsuperscript{1}{Department of Computer Science, University of Manchester, United Kingdom}\\
  \textsuperscript{2}{Digital Experimental Cancer Medicine Team}\\ {Cancer Research UK Manchester Institute, United Kingdom} \\
  \textsuperscript{3}{Idiap Research Institute, Switzerland} \\
  \texttt{\{zili.zhou, marco.valentino\}@manchester.ac.uk}\\ \texttt{donal.landers@digitalecmt.org} \\ \texttt{andre.freitas@idiap.ch} \\
}

\begin{document}

\maketitle

\begin{abstract}
This paper describes N-XKT (Neural encoding based on eXplanatory Knowledge Transfer), a novel method for the automatic transfer of explanatory knowledge through neural encoding mechanisms. We demonstrate that N-XKT is able to improve accuracy and generalization on science Question Answering (QA).
Specifically, by leveraging facts from background explanatory knowledge corpora, the N-XKT model shows a clear improvement on zero-shot QA. Furthermore, we show that N-XKT can be fine-tuned on a target QA dataset, enabling faster convergence and more accurate results. A systematic analysis is conducted to quantitatively analyze the performance of the N-XKT model and the impact of different categories of knowledge on the zero-shot generalization task.

\end{abstract}

\section{Introduction}

Contemporary Question Answering (QA) is evolving in the direction of addressing more abstractive reasoning tasks \cite{thayaparan2020survey,dua2019drop,clark2018think,OpenBookQA2018}, supported by multi-hop inference \cite{khot2019qasc,yang2018hotpotqa,banerjee-etal-2019-careful} and explanatory scientific facts \cite{jansen2019textgraphs,jansen2018WorldTree,jansen2016s}.

This trend of aiming to address more complex, multi-evidence and chained inference is pushing the envelope for novel representation and architectural patterns \cite{ding2019cognitive,qiu2019dynamically,asai2019learning,thayaparan2019identifying,kundu2019exploiting,valentino2021unification}, which are moving from modelling meaning from immediate distributional semantics patterns into deeper abstractive capabilities. 
This poses a paradigmatic challenge on the design of QA architectures, which need to operate over high-level semantic patterns and acquire the necessary knowledge to perform abstraction \cite{clark2018think}. At the same time, the design of new strategies to incorporate explanatory knowledge into neural representation has the potential to address fundamental data efficiency problems and promote zero-shot generalisation on out-of-distribution examples.

Explanation-based Science QA \cite{jansen2018WorldTree} provides a rich framework to evaluate these emerging requirements, as the task typically requires multi-hop reasoning through the composition of explanatory facts.
While existing approaches in the field mainly focus on the construction of natural language explanations
\cite{jansen2018WorldTree,jansen2019textgraphs}, this work aims to explore the impact of explanatory knowledge on zero-shot generalisation. 

In this paper, we argue that explanation-centred corpora can serve as a resource to boost zero-shot capabilities on Question Answering tasks which demand deeper inference. To this end, we explore the adoption of latent knowledge representations for supporting generalisation on downstream QA tasks requiring multi-hop inference.

Our hypothesis is that explanatory scientific knowledge expressed in natural language can be transferred into neural network representations, and subsequently used to achieve knowledge based inference on scientific QA tasks.
To validate this hypothesis, this paper proposes a \emph{unified} approach that frames Question Answering as an explanatory knowledge reasoning problem. The unification between the two tasks allows us to explore the adoption of pre-training strategies over explanatory knowledge bases, and subsequently leverage the same paradigm to generalise on the Question Answering task.

An empirical evaluation is performed on Transformers-based architectures adopting the WorldTree corpus as a knowledge base \cite{xie2020WorldTree,jansen2018WorldTree} and measuring generalisation on ARC \cite{clark2018think} and OpenbookQA \cite{OpenBookQA2018}. The main contributions of this paper are as follows:
\begin{itemize}
    \item We propose N-XKT, a neural mechanism for encoding and transferring explanatory knowledge for science QA. To the best of our knowledge, N-XKT is the first work tackling science QA tasks through the transfer of external explanatory knowledge via neural encoding mechanisms.

    \item We introduce the explanatory knowledge transfer task on explanation-centred knowledge bases, describing the methodology to implement N-XKT for knowledge acquisition and downstream Question Answering using Transformer-based models as neural encoders.
    
    \item We conduct a systematic empirical analysis to demonstrate the effectiveness of N-XKT on improving downstream QA accuracy and overall convergence speed in the training phase. An ablation analysis on different types of knowledge facts is performed to measure the impact of different knowledge categories.
    
    
    
    
    
    
\end{itemize}




\section{Related Work}


In this section we describe several works related to knowledge-based scientific QA.

\paragraph{Explanation Bank}

Explanation Bank\footnote{http://cognitiveai.org/explanationbank/} is a core component of the WorldTree corpus \cite{jansen2018WorldTree,xie2020WorldTree}. The dataset provides explanations for multiple-choice science questions in the form of graphs connecting questions and correct answers, where multiple sentences from a knowledge base (KB) are aggregated through lexical overlap between terms. The background knowledge used for the explanations is grouped in semi-structured tables, whose facts range from common-sense to core scientific statements. 
Explanation Bank has been proposed for the task of explanation regeneration \cite{jansen2019textgraphs} -- i.e. given a multiple-choice science question, regenerate the gold explanation supporting the correct answer. The explanation regeneration task has been framed as an Information Retrieval (IR) problem \cite{valentino2021unification}.
In this paper, we aim to leverage the knowledge expressed in the explanations to enhance generalisation and zero-shot capability on multiple-choice scientific question answering. 

\paragraph{Bidirectional Encoder Representations from Transformers}
BERT represents the foundation which defines the state-of-the-art in several NLP tasks \cite{devlin2018bert}. This model adopts a Transformer-based architecture composed of several layers of attention \cite{vaswani2017attention} that are used to learn a deep bidirectional representation of language. BERT-based models have demonstrated remarkable results in Question Answering when directly fine-tuned on the answer prediction task or additionally pre-trained using domain specific knowledge \cite{clark2019f,beltagy2019scibert}. A recent line of research attempts to enrich the input of BERT with background knowledge in the form of explanations in order to boost generalisation and accuracy for challenging QA settings. Here, the explanations are explicitly constructed through the adoption of language models \cite{rajani2019explain} or information retrieval (IR) approaches \cite{valentino2021unification,yadav2019quick}.
Conversely, this paper explores mechanisms to implicitly encode explanatory knowledge in the neural representation to improve the capability of performing downstream inference. Specifically, in this work, we adopt Transformers as text neural encoders.

\paragraph{Leveraging External Knowledge for Scientific QA}
Recently, many solutions have been proposed for science QA that leverage either external reference corpora \cite{Khot2017AnsweringCQ,Khashabi2018QuestionAA,Zhang2018KG2LT,banerjee-etal-2019-careful} or existing knowledge graphs \cite{banerjee-baral-2020-self,Li2015AnsweringES,Sachan2016ScienceQA,Wang2018YuanfudaoAS,Musa2019AnsweringSE,Zhong2019ImprovingQA}.
Generally, previous works rely on Information Retrieval models or on structural embeddings for Knowledge Bases, while our work focuses on directly encoding explanatory knowledge, evaluating it in a downstream scientific QA setting.

\begin{figure*}[t]
	\centering
	\includegraphics[width=\textwidth]{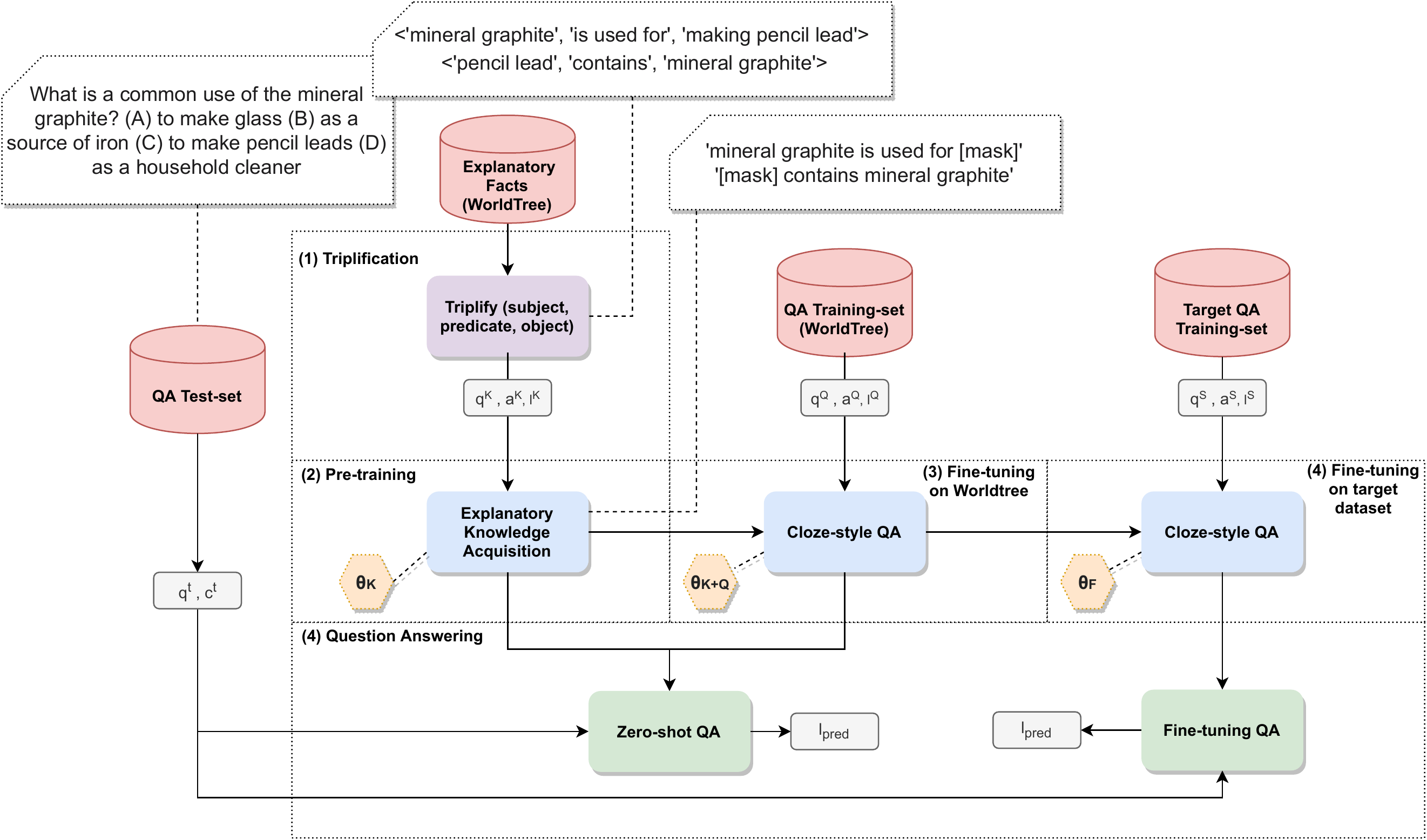}
	\caption[Approach Architecture]
	{Outline of the proposed approach.} 
	\label{fig:approach-architecture}
\end{figure*}

\section{Methodology}

Scientific Question Answering has the distinctive property of requiring the articulation of multi-hop and explanatory reasoning. This can be contrasted with the lexical-retrieval style of factoid Question Answering. Additionally, the explanatory chains required to arrive at the correct answer typically operate at an abstract level, through the combination of definitions and scientific laws \cite{thayaparan2020survey}. This characteristic makes the generalisation process more challenging, as the answer prediction model needs to acquire the ability to perform abstraction from the specific context in the question.

This paper hypothesises that it is possible to automatically transfer abstractive knowledge from explanatory facts into neural encoding representation for more accurate scientific QA, and for enabling zero-shot generalization.
To this end, we propose N-XKT (Neural encoding based on eXplanatory Knowledge Transfer) which encodes abstractive knowledge into neural representation to improve the effectiveness in both zero-shot QA task and fine-tuning based QA task.
The general neural encoding mechanism is evaluated adopting the following training tasks:
\begin{enumerate}

\item \textbf{Explanatory Knowledge Acquisition}:
In this pre-training task, the N-XKT model encodes the explanatory textual knowledge from a set of explanatory facts into supporting embeddings. This process aims to acquire the necessary explanatory knowledge to test generalization on downstream science QA. We frame this problem as a knowledge base completion task. Specifically, after casting each explanatory fact in the knowledge base into a tuple composed of subject, object, and predicate, the model is trained on completing each fact by alternatively masking each element in the tuple (additional details can be found in section \ref{sec:expl_knowledge}).

\item \textbf{Cloze-style Question Answering}:
To keep the encoding mechanism consistent with the pre-training explanatory knowledge acquisition task, we cast Multiple-choice Question Answering into a cloze-style QA problem. Specifically, we train the N-XKT model to complete the question with the expected candidate answer. This task aims to acquire additional knowledge for addressing downstream science QA since the patterns in the questions are typically more complex than the background explanatory facts (additional details can be found in section \ref{sec:questions_pattern}).
\end{enumerate}

The training tasks defined above can be used to encode different types and levels of knowledge into the N-XKT model, allowing us to perform a detailed evaluation on both zero-shot and fine-tuning-based Question Answering tasks.

Figure \ref{fig:approach-architecture} shows a schematic representation of the proposed approach.

\subsection{Explanatory Knowledge Acquisition}
\label{sec:expl_knowledge}
The WorldTree corpus \cite{jansen2018WorldTree} contains natural language explanatory facts, which are stored in semi-structured tables whose columns correspond to semantic roles. The knowledge base contains a total of 82 tables, where each table represents a different knowledge type, with different arity and argument types. N-XKT can be used as a unified approach for transferring knowledge from heterogeneous explanatory facts via a neural encoding mechanism.


To acquire the explanatory knowledge in a unified way for subsequent transfer learning, we normalize the semi-structured facts using a binary predicate-argument structure as typical practice in standard knowledge-base completion tasks \cite{bordes2013translating,wang2014knowledge,lin2015learning}. Specifically, for each table, we map the columns into three main components: subject, predicate, and object. After performing the mapping for each table in the knowledge base, we generate triples for all the facts in the knowledge base.

By framing the explanatory knowledge acquisition task as a knowledge base completion problem, we alternatively mask subjects and objects from the triples and train the model to predict the missing component in the triple by giving in input the remaining ones. Specifically, we simulate a question answering problem adopting either subject or object as an answer, and the other two components in the triple as a question.

The neural encoder of N-XKT learns an embedding representation for each pair in input. A softmax layer is added on top of the embedding to predict the probability of the missing component in the triple.
The configuration adopted for the N-XKT model is described in equation ~\ref{eq:likelihood-explanation-pre-training};.
\begin{equation}
	\begin{split}
	\theta_{K} \leftarrow \text{\text{argmin}}_\theta \mathcal{L}(\textbf{N-XKT}_{\theta}(q^{K},a^{K}),l^{K})
	\end{split}
	\label{eq:likelihood-explanation-pre-training}
\end{equation}
Here, $q^K$ and $a^K$ represent the simulated question-answer pair generated from a generic explanatory fact triple, while
$l^K$ represents the target labels (i.e. 1 if $a$ is the correct component for completing the triple, 0 otherwise).
$\theta_{K}$ is the set of parameters optimised during the explanatory knowledge acquisition stage.
The negative samples are generated by replacing each correct answer with a random component extracted from different explanatory facts in the knowledge base.

The transformer neural network is used as a textual neural encoder component of N-XKT, where
each question-answer pair is compiled into the input token sequence: 
\begin{equation}
[CLS] [question] [SEP] [answer] [SEP]
\label{eq:question-answer-input-compilation}
\end{equation}
The final hidden vector $C \in \mathbb{R}^H$ of the Transformer neural network that corresponds to the first input token ([CLS]) is used as an embedding to perform the final classification.

\subsection{Cloze-style Question Answering}
\label{sec:questions_pattern}
Normally, the explanatory knowledge patterns do not contain the complete information to address downstream Question Answering. However, the questions in WorldTree can be used as additional knowledge to deal with complex structured science questions, allowing N-XKT to learn to recognize more complex patterns.


To acquire additional knowledge while keeping the encoding mechanism consistent with the pre-training explanatory knowledge acquisition task, we cast Multiple-choice Question Answering into a cloze-style QA problem. The particular encoding configuration of the N-XKT model can be used in fact to address this type of question answering problems, where the model is trained to complete the question with the expected candidate answer. The detailed parameters and inputs adopted for cloze-style QA are described in equation \ref{eq:likelihood-QA-fine-tuning}:
\begin{equation}
	\begin{split}
	\theta_{K+Q} \leftarrow \text{argmin}_\theta \mathcal{L}(\textbf{N-XKT}_{\theta_{K}}(q^{Q},a^{Q}),l^{Q})
	\end{split}
	\label{eq:likelihood-QA-fine-tuning}
\end{equation}
The setting adopted for cloze-style QA is similar to the one adopted for explanatory knowledge acquisition, but with two main differences:
1) In this case, the question $q^Q$, the answer $a^Q$, and the target label $l^K$ are generated from the WorldTree multiple-choice question answering set, where the right candidate answer of each question acts as a positive sample, and the incorrect candidate answers act as the negative samples.
2) The initial parameters are initially set with $\theta_{K}$, that is, we adopt the parameters that have been optimised during the explanatory knowledge acquisition stage.

\subsection{Zero-shot and Fine-tuning Settings}

Given a multiple-choice science question, N-XKT can perform question answering by framing it as a sequence classification problem, where the question is paired with each candidate answer to compute a probability score. The candidate choice with highest score can then be selected as the predicted answer. We evaluate N-XKT in two different settings: zero-shot and fine-tuning-based QA.

Regarding the \textbf{zero-shot setting}, the N-XKT is trained only on the explanatory knowledge acquisition task and then directly tested on downstream Question Answering. We also evaluate the model trained jointly on explanatory knowledge and science questions in WorldTree, evaluating its generalization capabilities on different multiple-choice Question Answering datasets, such as ARC\footnote{\url{https://allenai.org/data/arc}} \cite{clark2018think} and OpenBook QA\footnote{\url{https://allenai.org/data/open-book-qa}} \cite{OpenBookQA2018}.
For each pair of question and candidate answer, the scores are computed as described in equation ~\ref{eq:prediction-zero-shot}. Here,
$(q^T,c^T)$ represent the test question and a candidate answer, while $l^T_{pred}$ is the score predicted by the model.
\begin{equation}
	\begin{split}
	l^T_{pred}=\textbf{N-XKT}_{\theta_{K+Q}}(q^T,c^T)
	\end{split}
	\label{eq:prediction-zero-shot}
\end{equation}

In the \textbf{fine-tuning setting}, the N-XKT model is additionally fine-tuned on each target QA dataset as in equation ~\ref{eq:prediction-fine-tuned}.
Here, $(q^{S},a^{S})$ represents a question-answer pair from the target QA training set, while $l^{S}$ is the label indicating whether the answer is correct or not.
\begin{equation}
	\begin{split}
	\theta_{F} \leftarrow \text{\text{argmin}}_\theta \mathcal{L}(\textbf{N-XKT}_{\theta_{K+Q}}(q^{S},a^{S}),l^{S})\\
	\end{split}
	\label{eq:prediction-fine-tuned}
\end{equation}
As shown in equation~\ref{eq:prediction-fine-tuned}, we adopt the same configuration as in the zero-shot setting, where the only difference is represented by the fine-tuned parameters set $\theta_{F}$:
\begin{equation}
	\begin{split}
	l^T_{pred}=\textbf{N-XKT}_{\theta_{F}}(q^T,c^T)
	\end{split}
	\label{eq:prediction-fine-tuned}
\end{equation}


\section{Empirical Evaluation}

We conduct our experiments on four widely used science QA datasets, WorldTree V2.0 \cite{xie2020WorldTree}, ARC Easy and Challenge \cite{clark2018think}, and Openbook QA \cite{OpenBookQA2018}.
The results tend to confirm our research hypothesis that explanatory knowledge encoding can improve generalization in downstream science Question Answering (QA) tasks.
Furthermore, we systematically analyze several factors which may have an impact on the final results, including the use of Transformer-based models with a larger number of parameters (BERT-large), testing the model on QA tasks using different types of explanatory background knowledge, and measuring training and test performance by further fine-tuning the model on other datasets.


\subsection{Experimental Setup}

\paragraph{QA dataset size.} In order to conduct a thorough quantitative analysis, we use four science QA datasets, WorldTree V2.0 \cite{xie2020WorldTree}, ARC Easy and Challenge \cite{clark2018think}, and Openbook QA \cite{OpenBookQA2018}.
The number of question-answer pairs in each dataset is listed in Table.~\ref{tab:question-answering-data-size}.

\begin{table}[t]
	\centering
	\caption{QA datasets size.}
	\label{tab:question-answering-data-size}
	\scalebox{0.9}{
		\begin{tabular}{@{}lccc@{}}
			\toprule
			Dataset & \#Train & \#Dev & \#Test \\
			\midrule
			WorldTree V2.0 & 3,947 & 1,019 & 4,165 \\
			ARC Easy & 2,251 & 570 & 2,376 \\
			ARC Challenge & 1,119 & 299 & 1,172 \\
			Openbook QA & 4,957 & 500 & 500 \\
			\bottomrule
		\end{tabular}
	}
\end{table}

\paragraph{Explanatory knowledge dataset size.} We encode different types of explanatory knowledge in the WorldTree corpus into Transformer neural networks. The statistics of the adopted explanatory facts are reported in Table~\ref{tab:explanatory-knowledge-data-size}.
Because we further analyze the impact of different types of knowledge, the number of each knowledge type is also given in the table.

\begin{table}[t]
	\centering
	\caption{Number of instances in each explanatory knowledge category.}
	\label{tab:explanatory-knowledge-data-size}
	\scalebox{0.9}{
		\begin{tabular}{@{}ll@{}}
			\toprule
			Type &  Size \\
			\midrule
			 All & 9,701 \\
			 Retrieval & 7,006 \\
			 Inference-supporting & 1,670 \\
			 Complex Inference & 1,025 \\
			\bottomrule
		\end{tabular}
	}
\end{table}

\begin{table*}[htbp]
    \small
	\centering
	\caption{N-XKT Question Answering accuracy results.}
	\label{tab:N-XKT-accuracy}
	\scalebox{1}{
		\begin{tabular}{@{}l|cc|cc|cc|cc@{}}
			\toprule
			 \multirow{2}{*}{Config}&  \multicolumn{2}{|c}{Explanation Bank} & \multicolumn{2}{|c}{ARC Easy} & \multicolumn{2}{|c}{ARC Challenge} & \multicolumn{2}{|c}{Openbook QA} \\
			 & Dev & Test & Dev & Test & Dev & Test & Dev & Test \\
			\midrule
			IR BM25 (K = 5) & 50.29\% & 44.55\% & 54.56\% & 50.00\% &  37.46\% & 31.14\% & 24.80\% & 26.80\% \\
			\midrule
			\multirow{1}{*}{K base} & 49.30\% & 44.74\% & 50.18\% & 50.89\% & 34.38\% & 33.17\% & 30.96\% & 32.72\% \\
			\multirow{1}{*}{Q base} & 44.86\% & 40.34\% & 50.81\% & 47.43\% & 24.41\% & 26.86\% & 27.92\% & 33.12\% \\
			\multirow{1}{*}{K+Q base} & 58.14\% & 50.42\% & 58.53\% & 57.98\% & 37.46\% & 35.87\% & 35.32\% & 37.60\% \\
			\midrule
			\multirow{1}{*}{K large} & 51.62\% & 45.85\% & 52.81\% & 52.58\% & 37.53\% & 33.07\% & 31.72\% & 34.12\% \\
			\multirow{1}{*}{Q large} & 47.54\% & 43.47\% & 53.61\% & 51.41\% & 27.09\% & 28.63\% & 28.24\% & 36.04\% \\
			\multirow{1}{*}{K+Q large} & \textbf{60.16\%} & \textbf{50.98}\% & \textbf{61.19}\% & 58.24\% & \textbf{39.00}\% & 37.63\% & 35.64\% & 38.20\% \\
			\midrule
			\multirow{1}{*}{base FT} & - & - & 53.61\% & 53.82\% & 36.72\% & 32.71\% & 53.64\% & 53.16\% \\
			\midrule
			\multirow{1}{*}{K base FT} & - & - & 53.61\% & 52.81\% & 35.79\% & 34.90\% & 53.60\% & 54.60\% \\
			\multirow{1}{*}{Q base FT} & - & - & 59.05\% & 58.44\% & 33.65\% & 35.09\% & 56.04\% & 57.08\% \\
			\multirow{1}{*}{K+Q base FT} & & & 59.33\% & \textbf{58.79\%} & 38.13\% & \textbf{38.09\%} & \textbf{56.12}\% & \textbf{56.56\%} \\
			\bottomrule
		\end{tabular}
	}
\end{table*}

\begin{table*}[t]
    \small
	\centering
	\caption{Question Answering accuracy results using different explanatory knowledge categories.}
	\label{tab:answer-selection-accuracy-expl-category}
	\scalebox{1}{
		\begin{tabular}{@{}l|l|cc|cc|cc|cc@{}}
			\toprule
			 \multirow{2}{*}{Knowledge} & \multirow{2}{*}{Config}&  \multicolumn{2}{|c}{Explanation Bank} & \multicolumn{2}{|c}{ARC Easy} & \multicolumn{2}{|c}{ARC Challenge} & \multicolumn{2}{|c}{Openbook QA} \\
			 & & Dev & Test & Dev & Test & Dev & Test & Dev & Test \\
			\midrule
			\multirow{1}{*}{None} & \multirow{1}{*}{Q base} & 44.86\% & 40.34\% & 50.81\% & 47.43\% & 24.41\% & 26.86\% & 27.92\% & 33.12\% \\
			\midrule
			\multirow{2}{*}{Retrieval} & \multirow{1}{*}{K base} & 39.05\% & 38.72\% & 44.42\% & 45.25\% & 23.75\% & 26.25\% & 27.12\% & 29.96\% \\
			& \multirow{1}{*}{K+Q base} & 51.00\% & 46.08\% & 51.79\% & 53.22\% & 34.65\% & 33.00\% & 31.96\% & 32.96\% \\
			\midrule
			\multirow{2}{*}{Inference-supporting} & \multirow{1}{*}{K base} & 41.60\% & 38.24\% & 45.96\% & 44.77\% & 26.09\% & 26.02\% & 27.40\% & 30.88\% \\
			& \multirow{1}{*}{K+Q base} & 52.72\% & 47.33\% & 54.35\% & 54.32\% & 34.85\% & 34.40\% & 33.64\% & 37.16\% \\
			\midrule
			\multirow{2}{*}{Complex inference} & \multirow{1}{*}{K base} & 41.01\% & 38.58\% & 46.32\% & 45.98\% & 24.95\% & 23.75\% & 26.96\% & 29.76\% \\
			& \multirow{1}{*}{K+Q base} & 52.99\% & 46.12\% & 55.30\% & 52.74\% & 34.78\% & 34.51\% & 32.08\% & 35.08\% \\
			\midrule
			\multirow{2}{*}{All} & \multirow{1}{*}{K base} & 49.30\% & 44.74\% & 50.18\% & 50.89\% & 34.38\% & 33.17\% & 30.96\% & 32.72\% \\
			& \multirow{1}{*}{K+Q base} & \textbf{58.14\%} & \textbf{50.42\%} & \textbf{58.53\%} & \textbf{57.98\%} & \textbf{37.46\%} & \textbf{35.87\%} & \textbf{35.32\%} & \textbf{37.60\%} \\
			\bottomrule
		\end{tabular}
	}
\end{table*}

\begin{table*}[htbp]
    \small
	\centering
	\caption{Accuracy comparison between N-XKT and othe approaches. External KB adopted by the models: 1.ARC-corpus \cite{clark2018think}, 2.ConceptNet \cite{Speer2017ConceptNet5A}, 3.Wikipedia (https://www.wikipedia.org/), 4.SciTail \cite{Khot2018SciTaiLAT}, 5.SNLI \cite{Bowman2015ALA}, 6.MultiNLI \cite{Williams2018ABC}, 7.RACE \cite{Lai2017RACELR}, 8.MCScript \cite{Ostermann2018MCScriptAN}, 9.WorldTree \cite{jansen2018WorldTree}t, 10. Synthetic Graph \cite{banerjee-baral-2020-self}.}
	\label{tab:answer-selection-accuracy}
	\scalebox{1}{
		\begin{tabular}{@{}lcccccc@{}}
			\toprule
			 & ARC Easy & ARC Challenge & Openbook QA & External KB & IR-based & Fine-tuned \\
			\midrule
            IR BM25 (K = 5) & 50.00\% & 31.14\% & 26.80\% & 9 & yes & no \\
            \citet{banerjee-baral-2020-self}&  31.80\% & 27.80\% & 34.40\% & 10 & no & no\\
			\citet{clark2018think} & 62.60\% & 20.30\% & - & 1 & yes & yes \\
			\citet{OpenBookQA2018} & - & - & 50.20\% & 2, 3 & yes & yes \\
			\citet{Khot2018SciTaiLAT} & 59.00\% & 27.10\% & 24.40\% & 4 & yes & yes \\
			\citet{Zhang2018KG2LT}  & - & 31.70\% & - &  1 & no & yes \\
			\citet{Yadav2018SanityCA} & 58.40\% & 26.60\% & - & none & no & yes \\
			\citet{Musa2019AnsweringSE} & 52.20\% & 33.20\% & - & 1 & yes & yes \\
			\citet{Zhong2019ImprovingQA} & - & 33.40\% & - & 2 & no & yes \\
			\citet{Pirtoaca2019ImprovingRQ} & 61.10\% & 26.90\% & - & 4, 5, 6 & no & yes \\
			\citet{Ni2019LearningTA} & - & 36.60\% & - & 7, 8 & no & yes \\
			$GPT^{II}$\cite{Radford2018ImprovingLU} & 57.00\%  & 38.20\% & 52.00\% & 7 & no & yes \\
			$RS^{II}$\cite{Sun2019ImprovingMR} & 66.60\% & 40.70\% & 55.20\% & 7 & no & yes \\
			\midrule
			N-XKT K+Q base (ours) & \textbf{57.98\%} & \textbf{35.87\%} & \textbf{37.60}\% & 9 & no & no \\
			\bottomrule
		\end{tabular}
	}
\end{table*}

\paragraph{Hyperparameters configuration.} We adjust two major hyperparameters for the training of the model, namely batch size and learning rate. We optimize the parameters considering the following combinations: we adopt training batch sizes in $\{16,32\}$, and learning rate in $\{1e-5,3e-5,5e-5\}$. The best results are obtained with batch size $32$ and learning rate $3e-5$ for the BERT-base model, and batch size $16$ and learning rate $1e-5$ for BERT-large \cite{devlin2018bert}.

\paragraph{Information Retrieval baseline.}
We adopt an Information Retrieval (IR) baseline similar to the one described in \citet{clark2018think}. Given a question $q$, for each candidate answer $c_i \in C = \{c_1, \ldots, c_n\}$, the IR solver uses BM25 vectors and cosine similarity to retrieve the top $K$ sentences in the WorldTree corpus that are most similar to the concatenation of $q$ and $c_i$. The score of a candidate answer $c_i$ is then obtained by considering the sum of the BM25 relevance scores associated to the retrieved sentences. The predicted answer corresponds to the candidate choice with the highest score. To test the generalisation of this approach on ARC and OpenbookQA, we keep the same background knowledge throughout the experiments.

\paragraph{Configuration Setting.}
We adopt different configurations in the experiments to control for  training data, Transformer model, and target QA test dataset fine-tuning. We report the different configurations in the ``Config'' column of Table~\ref{tab:N-XKT-accuracy} and Table~\ref{tab:answer-selection-accuracy-expl-category}. The label ``K'' indicates that the model is trained only on the explanatory knowledge acquisition task, ``Q'' means that the model is trained only on the cloze-style QA task using WorldTree as reference dataset, ``K+Q'' means that the model is pre-trained for explanatory knowledge acquisition and then further fine-tuned on cloze-style QA (again using only WorldTree as training dataset). Moreover, ``base'' means using BERT-base as Transformer model, while ``large'' means using BERT-large. Finally, ``FT'' means that the model is additionally fine-tuned on the target QA dataset's training data.

\begin{figure*}[t]
	\centering
	\includegraphics[width=\linewidth]{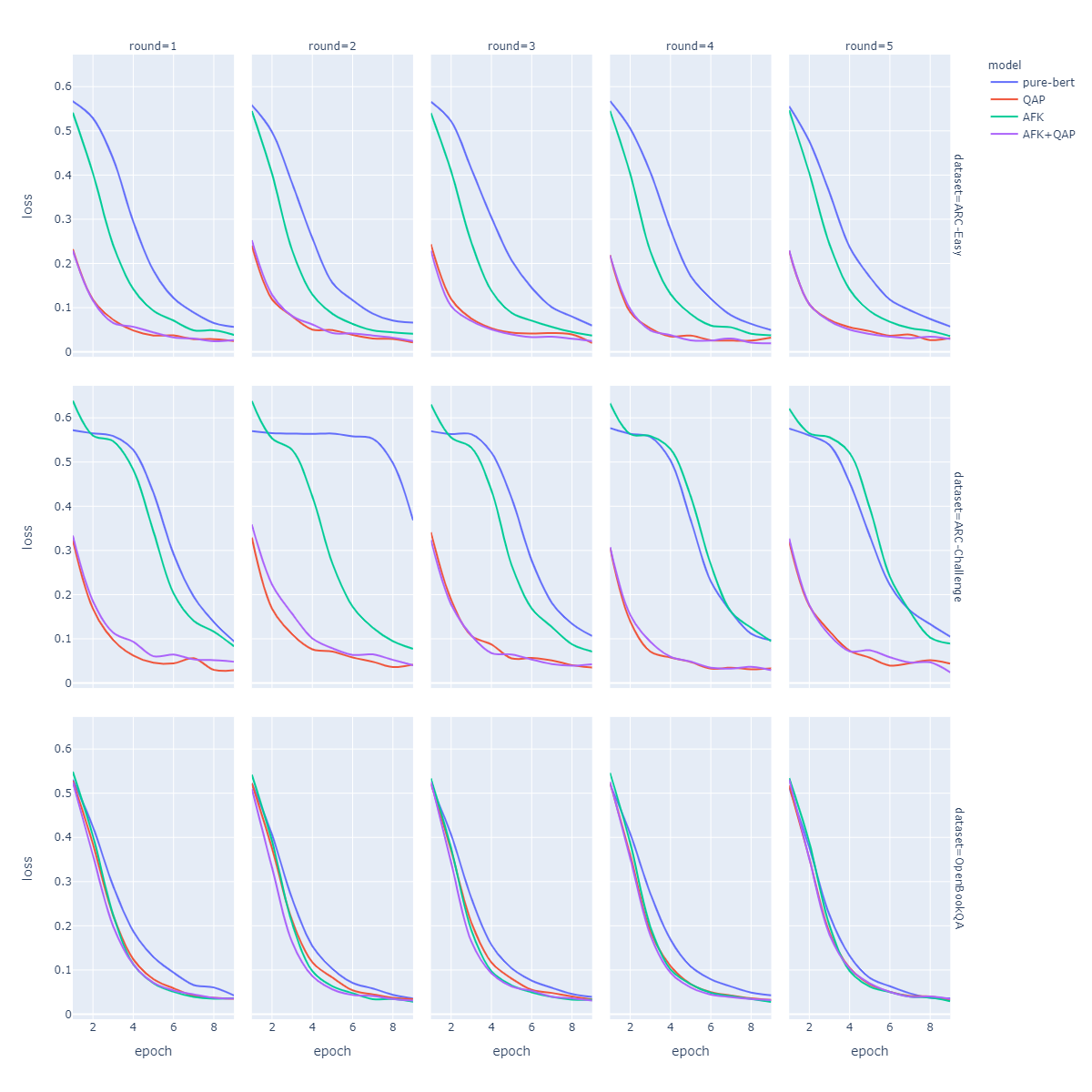}
	\caption[Fine-tuning covergence on other QA dataset]
	{Convergence curve when fine-tuning different version of N-XTK on the target QA datasets.}
	\label{fig:fine-tuning-covergence-other}
\end{figure*}

\subsection{Overall Results on Zero-shot Science Question Answering}

In Table ~\ref{tab:N-XKT-accuracy}, we report the performance of N-XKT under different configurations along with the accuracy of the BM25 baseline with $K = 5$ number of facts. The models are tested across multiple QA datasets including WorldTree, ARC, and OpenbookQA.

From the results, we derive the following conclusions. First, the proposed N-XKT model can clearly achieve better accuracy than the BM25 baseline since N-XKT uses Transformer-based neural mechanisms to acquire and encode external knowledge.
Second, using BERT-large instead of BERT-base as initial Transformer can improve the performance since BERT-large contains more parameters than BERT-base. However, we found that the advantage of using BERT-large is not significant since more parameters implies more resources needed for training.
Third,  we observe than N-XKT obtains better performance than pre-trained BERT when fine-tuning on the target datasets.

\subsection{Ablation Analysis on Impact of Different Explanatory Knowledge Types}

To understand the impact of different types of explanation on the final accuracy, we breakdown the facts stored in the knowledge base using three different categories (i.e., retrieval, inference-supporting and complex inference) and rerun the training of the N-XKT model using only one category per time.

The adopted categories are provided in the WorldTree corpus and can be described as follows:
\begin{itemize}
    \item \textit{Retrieval}: facts expressing knowledge about taxonomic relations and/or properties.
    \item \textit{Inference-Supporting}: Facts expressing knowledge about actions, affordances, requirements.
    \item \textit{Complex Inference}: Facts expressing knowledge about causality, processes, and if/then relationships.
\end{itemize}

The obtained accuracy is showed in Table~\ref{tab:answer-selection-accuracy-expl-category}. 
The results highlight the importance of using all the explanation categories to achieve the final accuracy for the combined approach. However, the retrieval category seems to have a higher impact on the generalisation. We believe that this result is due to the taxonomic knowledge encoded in the retrieval category (i.e. \emph{``x is a kind of y''}), which facilitates the acquisition of the implicit explanatory capabilities necessary for answering science questions.

In Table~\ref{tab:answer-selection-accuracy-expl-category}, we compare the impact of different explanatory knowledge types and get the following conclusion.
1) All three types of explanatory knowledge are helpful for further science QA task. The results using all three types of knowledge are significantly better than the results obtained when using no explanatory knowledge at all (first rown in Table \ref{tab:answer-selection-accuracy-expl-category}).
2) The model trained on all explanatory knowledge outperforms the models using each individual type of knowledge alone, confirming that different types of knowledge are complementary for achieving  the final performance.

\subsection{Evaluating Zero-shot N-XKT with Start-of-the-art baselines}

In Table~\ref{tab:answer-selection-accuracy}, we evaluate several start-of-the-art methods as baselines along with N-XKT trained only on the WorldTree. The table reports the accuracy results on ARC and OpenbookQA. In the ``External KB'' column, we list the external Knowledge Bases (KB) adopted by different models. The ``IR-based'' column indicates whether the model adopts Information Retrieval (IR) techniques, and the ``Fine-tuned'' column indicates whether the approach is fine-tuned on the target dataset.

Table~\ref{tab:answer-selection-accuracy} is intended to provide a general comparative analysis between N-XKT and the baseline models, most of them fine-tuned on the target datasets. N-XKT is able to achieve comparable performance under a transfer learning setting. The generalization performance of the proposed model is more noticeable for the ARC Challenge dataset, which requires the implicit encoding of more complex explanatory knowledge. 

\subsection{Improvement on Fine-tuning Convergence}

In Figure~\ref{fig:fine-tuning-covergence-other}, we visualize the convergence curve for the fine-tuning over three science QA tasks (ARC Easy, ARC Challenge and OpenBookQA), comparing a pure BERT-based N-XKT model with a pre-trained N-XKT models using different configurations, AFK (pre-trained on explanatory knowledge acquisition), QAP (pre-trained on WorldTree cloze-style QA), AFK+QAP (pre-trained on both). It is noticeable that the encoding of explanatory knowledge impacts the convergence of the model for all three datasets, with a particular emphasis on the two ARC variants.

\section{Conclusion}

In this paper, we proposed a neural encoding mechanism for explanatory knowledge acquisition and transfer, N-XKT. We evaluated the impact of the encoding mechanism on downstream science QA. The proposed model delivers better generalisation and accuracy for QA tasks that require multi-hop and explanatory inference. The proposed encoding mechanism can be used to deliver zero-shot inference capabilities, providing comparable performance when compared to supervised models on QA. These results supports the hypothesis that pre-training tasks targeting abstract and explanatory knowledge acquisition can constitute and important direction to improve inference capabilities and generalization of state-of-the-art neural models.

\bibliographystyle{acl_natbib}
\bibliography{anthology,acl2021}

\appendix
\section{Hyperparameters tuning}
The N-XKT mainly use a transformer network as natural language encoder component,
the hyperparameters of transformer network training have been tuned manually
for the optimisation is the maximisation of the accuracy in answer prediction.
Specifically, 3 parameters should be set for training, train batch size $\beta$, learning rate $\alpha$, and train epoch $\mathcal{N}$.
The values used in pre-training on explanation knowledge base are as follows:
\begin{itemize}
    \item $\beta$ = 32
    \item $\alpha$ = 5e-5
    \item $\mathcal{N}$ = 5
\end{itemize}
The values used in fine-tuning on Question Answer are as follows:
\begin{itemize}
    \item $\beta$ = 32
    \item $\alpha$ = 5e-5
    \item $\mathcal{N}$ = 40
\end{itemize}


\section{Data}
We use two versions of Explanation Bank Scientific Question Answer datasets in this paper.
The version 1 of Explanation Bank dataset can be downloaded at the following URL: \url{http://cognitiveai.org/dist/worldtree_corpus_textgraphs2019sharedtask_withgraphvis.zip}.
The version 2 of Explanation Bank dataset is available at the following URL: \url{https://github.com/cognitiveailab/tg2020task}.

\section{Computing Infrastructure}
To accelerate the training process of the experiments, we adopt a NVIDIA Tesla V100 GPU.

\section{Accuracy Results Including Standard Deviation}

\begin{table*}[htbp]
    \small
	\centering
	\caption{N-XKT Question Answering accuracy result comparison}
	\label{tab:N-XKT-accuracy}
	\scalebox{0.95}{
		\begin{tabular}{@{}l|cc|cc|cc|cc@{}}
			\toprule
			 \multirow{2}{*}{Config}&  \multicolumn{2}{|c}{Explanation Bank} & \multicolumn{2}{|c}{ARC Easy} & \multicolumn{2}{|c}{ARC Challenge} & \multicolumn{2}{|c}{Openbook QA} \\
			 & Dev & Test & Dev & Test & Dev & Test & Dev & Test \\
			\midrule
			IR BM25 (K = 5) & 50.29\% & 44.55\% & 54.56\% & 50.00\% &  37.46\% & 31.14\% & 24.80\% & 26.80\% \\
			\midrule
			\multirow{2}{*}{K base} & 49.30\% & 44.74\% & 50.18\% & 50.89\% & 34.38\% & 33.17\% & 30.96\% & 32.72\% \\
			& $\pm$0.0238 & $\pm$0.0166 & $\pm$0.0167 & $\pm$0.0198 & $\pm$0.0255 & $\pm$0.0165 & $\pm$0.0359 & $\pm$0.0273\\
			\multirow{2}{*}{Q base} & 44.86\% & 40.34\% & 50.81\% & 47.43\% & 24.41\% & 26.86\% & 27.92\% & 33.12\% \\
			& $\pm$0.0229 & $\pm$0.0087 & $\pm$0.0258 & $\pm$0.0136 & $\pm$0.0101 & $\pm$0.0049 & $\pm$0.0342 & $\pm$0.0176\\
			\multirow{2}{*}{K+Q base} & 58.14\% & \textbf{50.42\%} & 58.53\% & \textbf{57.98\%} & 37.46\% & \textbf{35.87\%} & 35.32\% & \textbf{37.60\%} \\
			& $\pm$0.0119 & \textbf{$\pm$0.0039} & $\pm$0.0047 & \textbf{$\pm$0.0014} & $\pm$0.0135 & \textbf{$\pm$0.0149} & $\pm$0.0124 & \textbf{$\pm$0.0085}\\
			\midrule
			\multirow{2}{*}{K large} & 51.62\% & 45.85\% & 52.81\% & 52.58\% & 37.53\% & 33.07\% & 31.72\% & 34.12\% \\
			& $\pm$0.0159 & $\pm$0.0089 & $\pm$0.004 & $\pm$0.0136 & $\pm$0.0109 & $\pm$0.0129 & $\pm$0.0199 & $\pm$0.0232\\
			\multirow{2}{*}{Q large} & 47.54\% & 43.47\% & 53.61\% & 51.41\% & 27.09\% & 28.63\% & 28.24\% & 36.04\% \\
			& $\pm$0.0131 & $\pm$0.0061 & $\pm$0.0176 & $\pm$0.0073 & $\pm$0.012 & $\pm$0.0125 & $\pm$0.0118 & $\pm$0.0167\\
			\multirow{2}{*}{K+Q large} & 60.16\% & 50.98\% & 61.19\% & 58.24\% & 39.00\% & 37.63\% & 35.64\% & 38.20\% \\
			& $\pm$0.0168 & $\pm$0.0102 & $\pm$0.0108 & $\pm$0.0076 & $\pm$0.0268 & $\pm$0.0155 & $\pm$0.0076 & $\pm$0.0161\\
			\midrule
			\multirow{2}{*}{base FT} & - & - & 53.61\% & 53.82\% & 36.72\% & 32.71\% & 53.64\% & 53.16\% \\
			& - & - & $\pm$0.0168 & $\pm$0.0093 & $\pm$0.0104 & $\pm$0.0086 & $\pm$0.0182 & $\pm$0.0223\\
			\midrule
			\multirow{2}{*}{K base FT} & - & - & 53.61\% & 52.81\% & 35.79\% & 34.90\% & 53.60\% & 54.60\% \\
			& - & - & $\pm$0.0159 & $\pm$0.0241 & $\pm$0.0218 & $\pm$0.0239 & $\pm$0.0248 & $\pm$0.0281\\
			\multirow{2}{*}{Q base FT} & - & - & 59.05\% & 58.44\% & 33.65\% & 35.09\% & 56.04\% & 57.08\% \\
			& - & - & $\pm$0.0177 & $\pm$0.0070 & $\pm$0.0280 & $\pm$0.0065 & $\pm$0.0126 & $\pm$0.0178\\
			\multirow{2}{*}{K+Q base FT} & & & 59.33\% & \textbf{58.79\%} & 38.13\% & \textbf{38.09\%} & 56.12\% & \textbf{56.56\%} \\
			& - & - & $\pm$0.0187 & \textbf{$\pm$0.0087} & $\pm$0.0224 & \textbf{$\pm$0.0124} & $\pm$0.0186 & \textbf{$\pm$0.0111}\\
			\bottomrule
		\end{tabular}
	}
\end{table*}

\begin{table*}[htbp]
    \small
	\centering
	\caption{Question Answering accuracy result in different abstractive knowledge categories}
	\label{tab:answer-selection-accuracy-expl-category}
	\scalebox{0.95}{
		\begin{tabular}{@{}l|l|cc|cc|cc|cc@{}}
			\toprule
			 \multirow{2}{*}{Knowledge} & \multirow{2}{*}{Config}&  \multicolumn{2}{|c}{Explanation Bank} & \multicolumn{2}{|c}{ARC Easy} & \multicolumn{2}{|c}{ARC Challenge} & \multicolumn{2}{|c}{Openbook QA} \\
			 & & Dev & Test & Dev & Test & Dev & Test & Dev & Test \\
			\midrule
			\multirow{2}{*}{None} & \multirow{2}{*}{Q base} & 44.86\% & 40.34\% & 50.81\% & 47.43\% & 24.41\% & 26.86\% & 27.92\% & 33.12\% \\
			& & $\pm$0.0229 & $\pm$0.0087 & $\pm$0.0258 & $\pm$0.0136 & $\pm$0.0101 & $\pm$0.0049 & $\pm$0.0342 & $\pm$0.0176\\
			\midrule
			\multirow{4}{*}{RET} & \multirow{2}{*}{K base} & 39.05\% & \textbf{38.72\%} & 44.42\% & \textbf{45.25\%} & 23.75\% & \textbf{26.25\%} & 27.12\% & \textbf{29.96\%} \\
			& & $\pm$0.0258 & \textbf{$\pm$0.0106} & $\pm$0.011 & \textbf{$\pm$0.0139} & $\pm$0.0165 & \textbf{$\pm$0.0141} & $\pm$0.0099 & \textbf{$\pm$0.0202} \\
			& \multirow{2}{*}{K+Q base} & 51.00\% & \textbf{46.08\%} & 51.79\% & \textbf{53.22\%} & 34.65\% & \textbf{33.00\%} & 31.96\% & \textbf{32.96\%} \\
			& & $\pm$0.0173 & \textbf{$\pm$0.0135} & $\pm$0.0178 & \textbf{$\pm$0.0141} & $\pm$0.0321 & \textbf{$\pm$0.0128} & $\pm$0.0192 & \textbf{$\pm$0.0182}\\
			\midrule
			\multirow{4}{*}{INSUPP} & \multirow{2}{*}{K base} & 41.60\% & \textbf{38.24\%} & 45.96\% & \textbf{44.77\%} & 26.09\% & \textbf{26.02\%} & 27.40\% & \textbf{30.88\%} \\
			& & $\pm$0.0149 & \textbf{$\pm$0.0075} & $\pm$0.0127 & \textbf{$\pm$0.0118} & $\pm$0.0164 & \textbf{$\pm$0.0099} & $\pm$0.0168 & \textbf{$\pm$0.0122}\\
			& \multirow{2}{*}{K+Q base} & 52.72\% & \textbf{47.33\%} & 54.35\% & \textbf{54.32\%} & 34.85\% & \textbf{34.40\%} & 33.64\% & \textbf{37.16\%} \\
			& & $\pm$0.0247 & \textbf{$\pm$0.0062} & $\pm$0.0206 & \textbf{$\pm$0.0092} & $\pm$0.031 & \textbf{$\pm$0.0128} & $\pm$0.0279 & \textbf{$\pm$0.0306}\\
			\midrule
			\multirow{4}{*}{COMPLEX} & \multirow{2}{*}{K base} & 41.01\% & \textbf{38.58\%} & 46.32\% & \textbf{45.98\%} & 24.95\% & \textbf{23.75\%} & 26.96\% & \textbf{29.76\%} \\
			& & $\pm$0.0132 & \textbf{$\pm$0.0035} & $\pm$0.0134 & \textbf{$\pm$0.0091} & $\pm$0.0263 & \textbf{$\pm$0.0066} & $\pm$0.012 & \textbf{$\pm$0.0163}\\
			& \multirow{2}{*}{K+Q base} & 52.99\% & \textbf{46.12\%} & 55.30\% & \textbf{52.74\%} & 34.78\% & \textbf{34.51\%} & 32.08\% & \textbf{35.08\%} \\
			& & $\pm$0.0098 & \textbf{$\pm$0.0131} & $\pm$0.0081 & \textbf{$\pm$0.0087} & $\pm$0.0112 & \textbf{$\pm$0.0194} & $\pm$0.018 & \textbf{$\pm$0.0153}\\
			\midrule
			\multirow{4}{*}{All} & \multirow{2}{*}{K base} & 49.30\% & 44.74\% & 50.18\% & 50.89\% & 34.38\% & 33.17\% & 30.96\% & 32.72\% \\
			& & $\pm$0.0238 & $\pm$0.0166 & $\pm$0.0167 & $\pm$0.0198 & $\pm$0.0255 & $\pm$0.0165 & $\pm$0.0359 & $\pm$0.0273\\
			& \multirow{2}{*}{K+Q base} & 58.14\% & 50.42\% & 58.53\% & 57.98\% & 37.46\% & 35.87\% & 35.32\% & 37.60\% \\
			& & $\pm$0.0119 & $\pm$0.0039 & $\pm$0.0047 & $\pm$0.0014 & $\pm$0.0135 & $\pm$0.0149 & $\pm$0.0124 & $\pm$0.0085\\
			\bottomrule
		\end{tabular}
	}
\end{table*}

We repeat the N-XKT model Question Answering training process on all the dataset for 5 times, each time with random parameters initialization.
Addition to the tables provided in paper, we report the detailed results with standard deviation in following tables.

Tab.~\ref{tab:N-XKT-accuracy} is for overall accuracy of N-XKT model on QA tasks, and Tab.~\ref{tab:answer-selection-accuracy-expl-category} is for ablation analysis results, only use part of explanations in training process.

\end{document}